\documentclass[twoside]{article}
\usepackage[accepted]{aistats2018}
\PassOptionsToPackage{hyphens}{url}

\usepackage{standalone}
\usepackage{graphicx}
\usepackage{float}
\usepackage{subfig}
\usepackage[authoryear, round]{natbib}
\usepackage{afterpage}
\usepackage{amssymb}
\usepackage{amsmath}
\usepackage{enumerate}
\usepackage{lscape}
\usepackage{setspace}
\usepackage[labelfont=bf]{caption}
\usepackage{sectsty}
\usepackage{fancyhdr}
\usepackage{dsfont}
\usepackage{bbm}
\usepackage{comment}
\usepackage{verbatim}
\usepackage{fancyvrb}
\usepackage{url}
\usepackage{enumitem}

\usepackage{dcolumn}

\usepackage{color}
\usepackage{bm}
\usepackage{amsfonts}
\usepackage{mathtools}
\usepackage{comment}
\usepackage{amsmath}
\usepackage{amssymb}
\usepackage{mathrsfs}
\usepackage{eqnarray}
\usepackage{enumitem}
\usepackage[ruled]{algorithm2e}
\usepackage[toc,page]{appendix}

\DeclareMathOperator*{\argmax}{argmax}

\setlist[enumerate]{itemsep=1mm}

\begin{document}

%

%

\twocolumn[

\aistatstitle{Gaussian Process Subset Scanning for Anomalous Pattern Detection in Non-iid Data}

\aistatsauthor{ William Herlands \And Edward McFowland III \And  Andrew G. Wilson \And Daniel B. Neill}

\aistatsaddress{ Carnegie Mellon University \And  University of Minnesota \And Cornell University \And Carnegie Mellon University } ]

\begin{abstract}\vspace{-2mm}
Identifying anomalous patterns in real-world data is essential for understanding where, when, and how systems deviate from their expected dynamics. Yet methods that separately consider the anomalousness of each individual data point have low detection power for subtle, emerging irregularities. Additionally, recent detection techniques based on subset scanning make strong independence assumptions and suffer degraded performance in correlated data. We introduce methods for identifying anomalous patterns in non-iid data by combining Gaussian processes with novel log-likelihood ratio statistic and subset scanning techniques. Our approaches are powerful, interpretable, and can integrate information across multiple data streams. We illustrate their performance on numeric simulations and three open source spatiotemporal datasets of opioid overdose deaths, 311 calls, and storm reports.
\end{abstract}

\vspace{-4mm}
\section{Introduction}
\label{sec:introduction}
\vspace{-1mm}

Anomalous pattern detection is the task of identifying subsets of data points that systematically differ from the underlying model. Identifying anomalous patterns in real-world data is critical for understanding how people and systems deviate from expected behavior. In the spatiotemporal domain, timely identification of such patterns can allow for effective interventions. For example, detecting anomalous increases in opioid deaths can enable health care workers to effectively target overdose prevention programs. Similarly, patterns of increased 311 calls can help cities to better target services and allocate resources.

To detect these anomalous patterns, we will address three key challenges. First, real-world data is extremely complex with non-trivial correlations across space, time, and other features. Treating data points as iid ignores  important covariance structure and will substantially overestimate the anomalousness of detected patterns. Second, an event of interest often affects multiple nearby points. Simply considering how anomalous is each individual point loses power to detect subtle anomalies. Third, anomalous patterns are often irregularly shaped or discontiguous due to latent demographic or geographic features. Searching for these complex patterns is important for precision and detection power, yet exhaustive methods are computationally intractable and may result in overfitting.

A sensible approach to this problem is model-based anomaly detection, where a distribution is fit to model ``regular'' data. Points with a low likelihood under this distribution are identified as anomalous \citep{chandola2009anomaly,hodge2004survey}. To address the complex correlations in real-world systems, Gaussian processes (GPs) provide a natural means of learning covariance structure from data. However, GP anomaly detection has been typically used to classify \textit{individual} points as outliers \citep{smith2014maritime,kowalska2012maritime,stegle2008gaussian}. Such approaches have difficulty when confronted with subtle anomalies, where each individual data point may seem to conform to the underlying distribution, yet when taken as a group, they form a collectively anomalous pattern. Thus anomalous pattern detection is a conceptually and statistically different problem than anomaly or outlier detection.

A few recent GP models consider anomalous intervals~\citep{reece2015anomaly} and sophisticated change points~\citep{saatcci2010gaussian,herlands2016scalable} to detect intervals of anomalous points. However, these methods (the first two of which are applied exclusively to one-dimensional data) are limited to contiguous intervals in the input domain and cannot model the irregularly shaped anomalies we expect in complex data. \cite{cheng2015video} recently developed an anomalous pattern detection technique for spatiotemporal data. However, this approach requires a corpus of anomaly-free training data, can only detect contiguous anomalous patterns, and is specific to video data.

In the statistics literature, spatial and subset scanning methods are commonly used to identify collectively anomalous subsets of data \citep{kulldorff1997spatial,neill2012fast}. By combining information across a subset of data elements, they generate a strong signal of anomalous behavior. These approaches compute a log-likelihood ratio (LLR) of subsets being drawn from a null or anomalous distribution. The LLR is a powerful statistic that measures how much evidence exists in the data to conclude if the subset exhibits abnormal behavior~\citep{kulldorff1997spatial,neill05}. A core challenge of subset scanning is searching through the $O(2^n)$ possible subsets of $n$ data elements~\citep{neill2004rapid, agarwal2006,duczmal07,wu2009}. \citet{neill2012fast} 
shows that certain LLR statistics satisfying a linear-time subset scanning (LTSS) property can be optimized in $O(n\log n)$ by ordering points according to a particular ``priority function'' and evaluating only $n$ of the $2^n$ subsets. However, LTSS assumes that we can compute the contributions of individual points to the LLR. This is possible only when assuming that data is uncorrelated under the null ~\citep{neill09-ebp}, 
yet when applied to non-iid data this independence assumption would result in substantial false positive rates, as correlated fluctuations will be mistaken for anomalous movements.

\vspace{-2mm}
\subsection{Contributions}
\vspace{-2mm}
In this paper we introduce novel techniques for identifying anomalous patterns in non-iid data. 
Our methods are powerful and interpretable. 
By combining naturally interpretable GPs with localized anomalous patterns we can describe the ``regular'' data dynamics as well as quantify and corroborate anomalous regions with domain experts. Our main contributions are:
\begin{enumerate}
\item Combining GP modeling with subset scanning for powerful and interpretable detection of anomalous patterns in highly correlated data.
\item Proposing a new likelihood ratio statistic and subset scan technique for correlated data that do not assume conditional independence.
\item Performing hold-out GP inference while computing our new likelihood ratio statistic conditioned on GP hyperparameters, to avoid corrupting the null model with anomalies.
\item Developing two novel, principled approaches to the NP-hard problem of searching for the most anomalous subset, through a new iterative method and an application of the Generalized Rayleigh Quotient respectively.
\item We demonstrate our methods on numeric simulations, opioid-related deaths, 311 calls for service data, and multiple streams of sewer flooding reports and tree damage reports, illustrating interpretable and policy-relevant results.
\end{enumerate}

The paper proceeds as follows: \S\ref{sec:GP} provides background on GPs. \S\ref{sec:LLRiid} introduces a novel log-likelihood ratio statistic for non-iid data. \S\ref{sec:scanning} details the Gaussian Process Neighborhood Scan (GPNS) and the Gaussian Process Subset Scan (GPSS). Experimental results on numerical and real data are presented in \S\ref{sec:exp}.

\vspace{-1mm}
\section{Gaussian Processes}
\label{sec:GP}
\vspace{-1mm}
Consider data, $(x,y)$, where $x = \{x_1,\dots,x_n\}, x_i \in \mathbb{R}^D$, are inputs or covariates, and $y = \{y_1,\dots,y_n\}$, $y_i \in \mathbb{R}$ are outputs or response variables indexed by $x$. We assume that $y$ is generated from $x$ by a latent function with a GP prior. In particular, $y = f(x) + \epsilon$, $f(x) \sim \mathcal{GP}(m(x), k(x,x'))$

A GP is a nonparametric prior over functions completely specified by mean and covariance functions. The mean function, $m(x) = \mathbb{E}[f(x)]$, is the prior expectation of $f(x)$. The covariance function is given by $k(x,x') = \text{cov}(f(x), f(x'))$ ~\citep{rasmussen2006gaussian}.

In this paper we use three important properties of GPs. First, we can draw samples from a GP prior since conditional on GP hyperparameters any finite collection of function values is distributed $\mathcal{N}(m(x), k(x,x))$

Second, if a function has a Gaussian noise model, $\epsilon \sim \mathcal{N}(0, \sigma_\epsilon^2)$, then conditional on hyperparameters and data $(x,y)$, we can derive a closed form expression for the predictive distribution of $f(x^*)$,
\begin{equation}
\label{eq:GP_posterior}
\begin{aligned}
&f(x^*) | x,y,x^* \sim \mathcal{N}\Big( k(x^*, x) [k(x,x) + \sigma_\epsilon^2I]^{-1}y,\\
&k(x^*,x^*) - k(x^*,x) [k(x,x) + \sigma_\epsilon^2I]^{-1} k(x,x^*) \Big)
\end{aligned}
\end{equation}

Third, GP hyperparameters, $\theta$, can be learned by maximum likelihood optimization. While naively this requires $O(n^3)$ computations, we use scalable GP learning in the structured kernel inference (SKI) framework~\citep{wilson2015kernel} for $O(n)$ scalability.

\vspace{-1mm}
\section{LLR statistic for non-iid data}
\label{sec:LLRiid}
\vspace{-2mm}
Considering $(x,y)$ as defined in \S\ref{sec:GP}, we are interested in anomalous patterns that systematically differ from the underlying data distribution. We frame this search as an LLR comparison between a null model of ``regular'' behavior and an alternative model of ``anomalous'' behavior. A single latent GP defines both models. Subsets of data with the highest LLR scores are identified as the most anomalous.

Using a GP as the foundational modeling technique enables us to learn complex covariance structure and seamlessly extend to high dimensions as well as missing data. GPs are also naturally interpretable, which can provide insight about the ``regular'' data dynamics.

Consider a given subset of data points defined by the binary weighting vector $w$, where $w_i = 1$ if $(x_i,y_i)$ is included in the subset and $w_i = 0$ if excluded.  Our null model, $H_0$, assumes that all points (regardless of $w_i$) are drawn from a function with a GP prior: $y = f(x) + \epsilon$, where $f(x) \sim GP(\theta_0)$ and $\epsilon \sim \mathcal{N}(0,\sigma_\epsilon^2 I)$.  Our alternative model, $H_1(w)$, assumes that $y_i = f(x_i) + \epsilon$ for $w_i = 0$, and $y_i = g(f(x_i),\theta_1) + \epsilon$ for $w_i = 1$,  where $g(\cdot)$ is any function of the latent GP. 

Here we focus on the case of a mean shift, $g(f(x), \theta_1) = f(x) +\beta$, $\beta \in \mathbb{R}^1$. The covariance structure remains the same in the null and alternative models. This allows us to efficiently compute the posterior mean vector $\mu$ and covariance matrix $\Sigma$ through GP inference, where $y \sim \mathcal{N}(\mu,\Sigma)$ under $H_0$, and $y \sim \mathcal{N}(\mu+\beta w,\Sigma)$ under $H_1(w)$. For posterior $\mu$ and $\Sigma$ we condition on all data outside the subset of points represented by $w$, ensuring that null model estimates are not corrupted by anomalous observations.  However, since anomalies are assumed to be rare, their influence on parameter estimation is minimal. Therefore we use all $(x,y)$ for GP learning of the parameters of the null model $\theta_0$. 

We concentrate on mean changes since many real world cases concern anomalous levels of a quantity. Increases in localized drug overdoses, crime, and calls for city service are all mean shifts of great importance. Methods for identifying arbitrary changes in distribution -- while able to detect other sorts of patterns -- have reduced power to detect such mean shifts, due to more diffuse inductive biases. Persistent changes in covariance structure are typically considered changepoints and require substantial data in both regimes as opposed to the localized anomalous patterns we detect.

To measure how anomalous is a subset defined by $w$, we compute the generalized log-likelihood ratio, $LLR(w) = \max_\beta LLR(w \:|\: \beta)$, where:
\vspace{-2mm}
\begin{equation}
\begin{aligned}
&LLR(w \:|\: \beta) = \log \frac{\texttt{MNPDF}(y-\beta w \:|\: \mu, \Sigma)}{\texttt{MNPDF}(y \:|\: \mu, \Sigma)}
\end{aligned}
\end{equation}
\vspace{-2mm}
Here \texttt{MNPDF} is the multivariate normal probability density function. The most anomalous subset, $w^*$, is
\begin{equation}
\begin{aligned}
\label{eq:LLR_neigh}
w^* &= \argmax_w LLR(w) \\
&= \argmax_{w} \max_\beta -\frac{\beta^2}{2}w^T E w + \beta w^T  E(y - \mu)
\end{aligned}
\end{equation}
where $E=\Sigma^{-1}$ for notational brevity. Conditional on $w$, the MLE $\beta^* = \arg\max_\beta LLR(w)$ can be calculated in closed form, $\beta^* = [w^T  E(y - \mu)]/[w^T E w ]$ (see Appendix~\ref{sec:appendix_MLE}).  Nevertheless, maximizing $LLR(w)$ is an NP-complete Integer Quadratic Program~\citep{del2014mixed}, so an optimal solution requires exponential-time computation. Note that the LTSS condition for a log linear-time subset search described in \citet{neill2012fast} does not apply, since it requires independent data with a diagonal covariance matrix.


\vspace{-1mm}
\subsection{Randomization testing}
\vspace{-1mm}
Given a method for finding anomalous subsets, the following randomization testing procedure determines an $\alpha$-level significance threshold for $LLR(w)$ conditional on the parameters of the null model: 
\vspace{-0.2cm}
\begin{enumerate}
\item Repeatedly draw $y^{(r)} \sim GP(\theta_0)$, at the same covariates, $x$, as the real data for $r=1...R$.
\item Scan over $(x,y^{(r)})$ with the chosen subset searching method. For each randomization $r$ save the most anomalous LLR value, $LLR(w^{*,(r)})$.
\item Determine an $\alpha$-level threshold for significance based on the $(1-\alpha)$ quantile of the $R$ maximum LLR values, above which any $LLR(w)$ from the original scan is considered statistically significant.
\end{enumerate}

\vspace{-4mm}
\section{Efficient subset scanning}
\label{sec:scanning}
\vspace{-1mm}
Having defined the LLR scan statistic to evaluate how anomalous is a given subset, we must now decide over which subsets to scan. Unconstrained optimization over $O(2^n)$ subsets is computationally infeasible for an exhaustive search. Additionally, an unconstrained search may return an unrelated set of points, reducing interpretability and increasing the potential for overfitting.  Anomalous events in human data, such as drug usage and requests for government services, often affect multiple nearby points. Thus we assume that anomalous points are near one another. For example, in spatiotemporal data we assume that anomalous points are clustered in space and time. Following \citet{neill2012fast}, we define the local ``$k$-neighborhood'' of each data point, consisting of that point and its $k-1$ nearest neighbors, for some $k$. We propose two approaches for using these neighborhoods to identify anomalous patterns: Gaussian Process Neighborhood Scan (GPNS) and Gaussian Process Subset Scan (GPSS).

\vspace{-2mm}
\subsection{GP Neighborhood Scan (GPNS)}
\vspace{-1mm}
Given a maximum neighborhood size $k_{max}$, GPNS searches over the $O(nk_{max})$ local neighborhoods consisting of the $k$-neighborhood for each point where $k = \{1, 2, \ldots, k_{max}\}$. 
Where neighborhoods are defined by Euclidean distance, the set of search regions are circular in shape. For each neighborhood, $(x^{(n)},y^{(n)})$, we obtain posterior $\mu$ and $\Sigma$ conditional on $\theta_0$ and points $(x^{(-n)},y^{(-n)})$. We then compute $LLR(w)$ for the neighborhood where $w=\vec{1}$, i.e., we evaluate the alternative hypothesis of the entire neighborhood being anomalous. GPNS pseudocode is presented in Alg.~\ref{alg:GPNS}.
\vspace{-3mm}
\begin{algorithm}[h]
\SetAlgoLined
  \caption{GPNS}
        \For{$k=1:k_{max}$}{
            \For{$(x_i,y_i)$, $i=1:n$}{
                Define $k$-neighborhood, $n^{(k,i)}$, and infer ($\mu,\Sigma$)\;
                Set $w^{(k,i)}=\vec{1} \in \{0,1\}^{k}$\;
                Compute $\beta^*$ given $w^{(k,i)}$\;
                Compute $LLR(w^{(k,i)})$\;
            }
        }
        Choose $n^* = \argmax_{n^{(k,i)}} LLR_{n^{(k,i)}}$\;
        Randomization testing for significance\;
	\label{alg:GPNS}
\end{algorithm}

\vspace{-3mm}
\subsection{GP Subset Scan (GPSS)}
\vspace{-1mm}
While GPNS simplifies the exponential search, it requires constraining assumptions about the shape of neighborhoods and is only able to discover contiguous, spherical anomalous patterns. While there are approaches to increase the variety of neighborhood shapes without substantially degrading computational efficiency \citep{kulldorff2006elliptic,neill2004rapid,kulldorff2001prospective}, these methods still require strict specification of potential anomalies. Such foreknowledge is unrealistic in real-world applications where natural boundaries, demographics, and stochastic effects lead to irregularly-shaped patterns. In such cases GPNS has reduced detection and explanatory power.

To flexibly detect irregularly-shaped patterns, GPSS conducts an unconstrained search for the most anomalous subset within neighborhoods of fixed size $k$. Specifically, we identify the subset of points $(x^{(s)},y^{(s)}) \subseteq (x^{(n)},y^{(n)})$ that maximize the LLR within each neighborhood. This allows us to identify highly irregular and even non-contiguous anomalous patterns. By restricting the search within a local neighborhood, we ensure that the identified patterns are coherent and interpretable. GPSS requires evaluating $O(n)$ neighborhoods, as presented in Alg.~\ref{alg:GPSS}.
\vspace{-1mm}
\begin{algorithm}[h]
\SetAlgoLined
    \caption{GPSS}
        Fix $k$ at some size\;
        \For{$(x_i,y_i)$, $i=1:n$}{
            Define $k$-neighborhood, $n^{(i)}$, and infer ($\mu$,$\Sigma$)\;
            Approximate the optimal subset, $s^{(i)} \subseteq n^{(i)}$\;
            Set each $w_j^{(i)}=1(j\in s^{(i)})$\;
            Compute $\beta^*$ given $w^{(i)}$\;
            Compute $LLR(w^{(i)})$\;
        }
        Choose $s^* = \argmax_{s^{(i)}} LLR_{s^{(i)}}$\;
        Randomization testing for significance\;
  \label{alg:GPSS}
\end{algorithm}
\vspace{-3mm}

Unfortunately, this procedure requires finding $w \in \{0,1\}^k$ that maximizes the LLR of a subset within the neighborhood, $\argmax_{w} \text{-}\frac{1}{2}w^T\beta E w\beta + w^T\beta  E(y^{(n)} \text{-} \mu)$. This is still an Integer Quadratic Program, whose optimal solution is intractable even for moderately sized neighborhoods. Instead, below we formulate three approaches for finding approximate solutions.

\vspace{-1mm}
\subsubsection{$\beta_{MAX}$ for conditionally optimal subset}
\vspace{-1mm}

Due to the full rank covariance matrix, we are unable to disentangle the individual contributions from each point to the LLR. However, if we condition on some subset of points, $w$, we are able to compute the conditional contribution of each point. First, note that conditional on $w$ we can decompose $w^*$ from Eq.~\ref{eq:LLR_neigh} into a sum over each of the $m$ points in the neighborhood
\begin{equation}
\label{eq:LLR_neigh_decompose}
\begin{aligned}
&w^T\beta  E(y^{(n)}- \mu)-\frac{1}{2}w^T\beta E w\beta \\
&= \sum_i w_i\Big[ \beta \big( E(y^{(n)}- \mu)\big)_i-\frac{1}{2}\big(\sum_{j\neq i} w_j E_{j,i} + E_{i,i}\big) \beta^2   \Big]
\end{aligned}
\end{equation}
The contribution of point $(x_i,y_i)$ to the LLR is the difference in LLR between $w_i=0$ and $w_i=1$. Due to the outer and inner sums, the change in the LLR is:
\begin{equation}
\label{eq:LLR_neigh_contribute}
\begin{aligned}
 \beta \big( E(y^{(n)}- \mu)\big)_i-\frac{1}{2}\big(\sum_{j\neq i} 2w_j E_{j,i} + E_{i,i}\big) \beta^2 
\end{aligned}
\end{equation}
To maximize the LLR a point is only added to the subset if its contribution is positive. By setting Eq.~\ref{eq:LLR_neigh_contribute} to zero we can compute $\beta_{MAX_i}$, the maximum $\beta$ value for which to include point $(x_i,y_i)$.
\begin{equation}
\begin{aligned}
\beta_{MAX_i} = \Big[2\big( E(y^{(n)}\text{-} \mu)\big)_i\Big]/\Big[\sum_{j\neq i} 2w_j E_{j,i} + E_{i,i}\Big]
\end{aligned}
\end{equation}
As proved in \citet{speakman2016penalized}, we obtain the conditional optimal subset by using $\beta_{MAX}$ as a priority function, ranking each data point by $\beta_{MAX_i}$, and iteratively compute the score function for subsets including each additional point. This yields a log linear search over data points. Such an approach identifies the most anomalous subset with a positive mean shift. To find the most anomalous subset with a negative mean shift we simply rank data points by $-\beta_{MAX_i}$ 

Since the derivation of $\beta_{MAX}$ is conditional on a subset $w$, we obtain the \textit{conditional} optimal subset. In order to approximate an optimal solution we iteratively compute the conditional optimal subset beginning with a null subset, $w=\vec{0}$. This requires $O(\ell k \log(k))$ computation for some $\ell$ number of iterations. Pseudo-code for this algorithm can be found in Appendix~\ref{sec:appendix_iterative}.

For a diagonal $\Sigma$, $\beta_{MAX}$ orders points according to $2(y^{(n)}_i-\mu_i)$, which is equivalent to the LTSS priority function for an independent Gaussian subset scan~\citep{speakman2016penalized}.  Thus $\beta_{MAX}$ approach identifies the optimal subset in the independent case and is conditionally optimal in the dependent case.

\vspace{-2mm}
\subsubsection{Generalized Rayleigh Quotient method}
\vspace{-1mm}

We consider an alternative optimization approach to obtain an approximately optimal subset. Consider plugging the MLE solution, $\beta^*$, into $w^*$ from Eq.~\ref{eq:LLR_neigh},
\begin{equation}
\label{eq:GRQ_derive}
\begin{aligned}
w^* = \argmax_{w} \Big[w^T \big( E(y^{(n)}\text{-}\mu) (y^{(n)}\text{-}\mu)^T   E\big) w \Big] / \\
 \Big[w^T\big(2  E\big) w \Big]
\end{aligned}
\end{equation}
If we relax $w$ such that $w\in \mathbb{R}^m$, this can be re-written as the generalized Rayleigh quotient, $(w^T A w)/(w^T B w)$,
where $A = E (y^{(n)}-\mu) (y^{(n)}-\mu)^T E$, and $B = 2 E$. Note that $A$ is a symmetric matrix and $B$ is a Hermitian positive-definite matrix. Taking the Cholesky decomposition $B=LL^T$, the generalized Rayleigh quotient can be written as a Rayleigh quotient~\citep{yu2013kernel}, $R(A',w') = (w'^T A' w')/(w'^T w')$, where $A' = L^{-1}A L^{T^{-1}} $ and $w'=L^T w$. The maximum $w'$ of the Rayleigh quotient, $w'_{max} = \argmax_{w'} R(A',w') = \argmax_{w'} (w'^T A w')/(w'^T w') = v^{(max)}$,
is the largest eigenvector of $A'$. Since we defined $w'=L^T w$, then the maximum $w_{max}=L^{T^{-1}}v^{(max)}$ is the relaxed solution to our original optimization problem from Eq.~\ref{eq:GRQ_derive}.

Although $w_{max}$ has non-integer elements, the ordering of the elements of this eigenvector corresponds to the importance of the data points in the neighborhood. Thus we scan over the ordered elements of $w_{max}$, iteratively adding each to the subset. Maximizing $LLR(w)$ over this linear number of subsets provides an approximate solution to the constrained integer program. 

\vspace{-1mm}
\subsubsection{Forward stepwise optimization}
\vspace{-1mm}

A third approximation approach uses a greedy forward stepwise algorithm that iteratively sets one element $w_i=1$ such that the objective is minimized in each iteration. Once the objective cannot be further minimized the optimization is terminated, thereby providing a greedy optimal solution. 
For a neighborhood of size $k$, the stepwise approach may require up to $k$ iterations, evaluating $O(k)$ subsets at each iteration for a total of $O(k^2)$ computations.

\vspace{-1mm}
\subsection{Efficient Multi-Stream Search}
\label{sec:mutli-stream}
\vspace{-1mm}

Often we are interested in searching for anomalous patterns across multiple dimensions, or streams, of data. For example, anomalous patterns of damaged trees and sewer flooding can help localize severe storm damage. Multi-stream search can enhance the signal of subtle anomalies that affect multiple streams, and  reduce false positive detections when perturbations in a single stream are not important to the application.

In principle, GPNS and GPSS can handle multiple streams by stacking the data from each stream and adding a final dimension to indicate from which stream the data came. Yet naive GP inference requires $O(n^3)$ complexity, so repeatedly concatenating data from multiple streams quickly leads to scalability issues. On the other hand, Kronecker-based scalability  require a kernel that is multiplicatively decomposable over the input dimensions~\citep{saatcci2012scalable}. This implies that the prior correlation structure is the same over all data dimensions except for the stream indicator. For example, Kronecker structure in spatiotemporal settings constrains streams to have the same prior spatiotemporal correlations. This assumption is overly restrictive for the complex data in which we are interested.

Instead, we learn independent GPs for each stream of data and then scan over neighborhoods in the data jointly for all streams. Posteriors for each stream are independently inferred from the associated GP. Thus for streams $s=1,...,S$, the posterior distribution for subset scanning contains a block diagonal covariance,
\[
\mathcal{N}\left(
\begin{bmatrix}
    \mu_{1}\\
    \vdots\\
    \mu_{S}
\end{bmatrix},
\begin{bmatrix}
    \Sigma_{1}  &0 &0 \\
    0 &  \ddots & 0 \\
    0 & 0 &\Sigma_S
\end{bmatrix}
\right)
\]
In this manner each stream can flexibly learn different prior covariance structures while still ensuring scalability equivalent to single-stream GPNS and GPSS. The one drawback of this  approach is that inter-stream covariance information is not exploited for GP inference.

\vspace{-1mm}
\section{Experiments}
\vspace{-1mm}
\label{sec:exp}

We evaluate GPNS and GPSS using numeric simulations and three urban spatiotemporal datasets. We compare the methods against a number of competitive baseline algorithms from contemporary literature. First, we compare to an independent Gaussian subset scan, a state of the art anomalous pattern detection algorithm~\citep{neill09-ebp,neill2012fast}. Additionally, we compare against a standard GP anomaly detection approach~\citep{kowalska2012maritime,stegle2008gaussian}, in which we use the posterior distribution of the null GP model $\theta_0$ regressed over the entire dataset to classify points beyond a given level-$\alpha$ significance threshold as anomalies. While all GP methods in this paper are agnostic to kernel choice, an RBF kernel and linear mean function were used for all experiments.

Although anomalous pattern detection is a distinct problem from outlier or anomalous point detection, we also compare against two commonly used outlier detection techniques: a one-class SVM~\citep{scholkopf2001estimating} and 
robust multivariate outlier detection using the Mahalanobis distance~\citep{rousseeuw2005robust,rousseeuw1990unmasking}. 

An additional real data analysis is in Appendix~\ref{sec:exp_school}.

\vspace{-1mm}
\subsection{Numeric experiments}
\label{sec:exp_numeric}
\vspace{-1mm}

For each numeric test, baseline data is drawn from a 2D GP~\citep{GPML40}. Multiplicative anomalies of arbitrary shape are injected by scaling randomly sampled points, within a randomly chosen neighborhood, by a factor of $\geq1$. (Note that this simulation does not correspond to our method's assumption of an additive mean shift.) The most anomalous subset is computed using GPSS methods and baseline approaches. For the baseline GP approach and one-class SVM we provide additional information (the true percentage of the anomalous data) in order to determine their threshold levels. 
\begin{figure}[h]
\center
\hspace*{-0.4cm}
\includegraphics[width=1.1\linewidth]{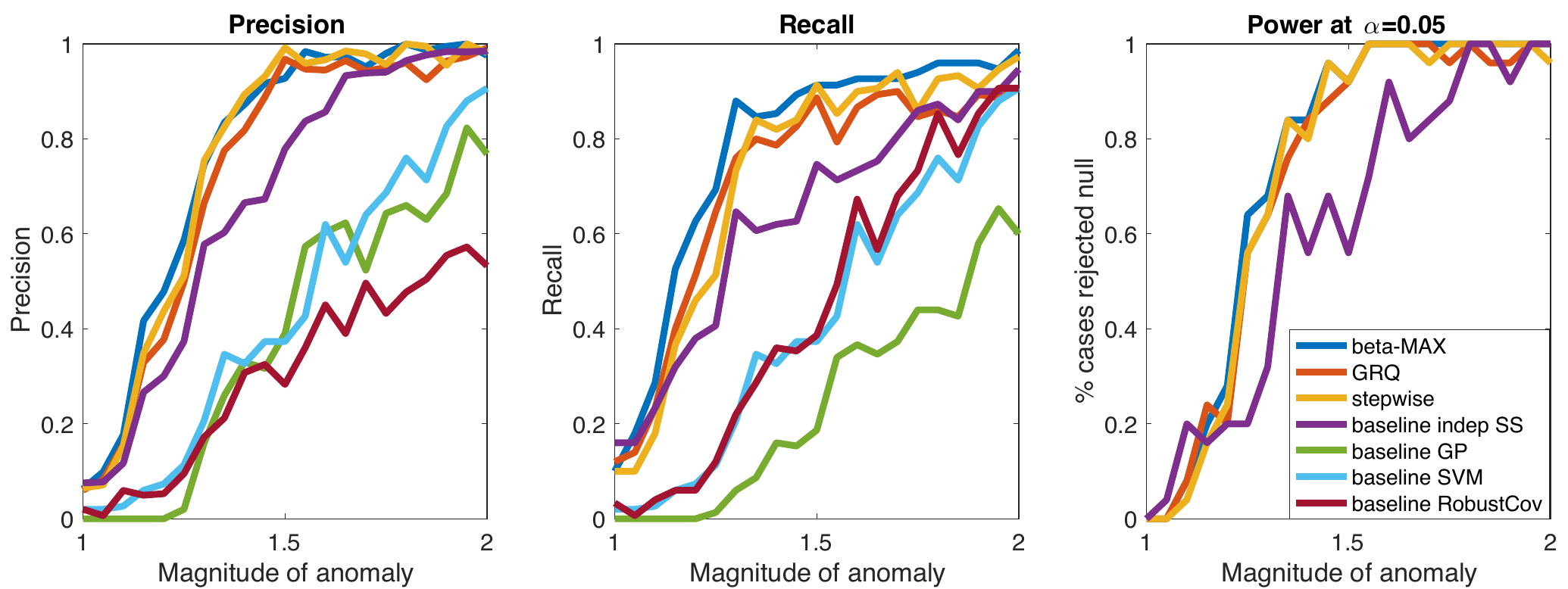}
\caption{Precision, recall, and power at $\alpha=0.05$ for GPSS methods and baseline anomaly detection approaches. The three GPSS methods dominate in all cases with the $\beta_{MAX}$ performing best overall.}
\label{fig:run_synthetic_results}
\vspace{-0.25cm}
\end{figure}

Varying the multiplicative factor between 1 and 2 we compute the average precision and recall in Fig.~\ref{fig:run_synthetic_results} over 50 tests in a 400 point grid for each multiplicative factor. Randomization testing ($\alpha = .05$) is performed for each synthetic test to determine the score threshold for significance. For precision and recall, truly anomalous points are ``positive'' and all other data is ``negative.'' The GPSS approaches dominate all other methods for nearly the entire test range, with $\beta_{MAX}$ performing best overall.

\begin{figure}[h]
  \begin{minipage}[b]{0.50\linewidth}
    \centering
    \hspace*{-0.2cm} 
    \includegraphics[width=1.1\linewidth]{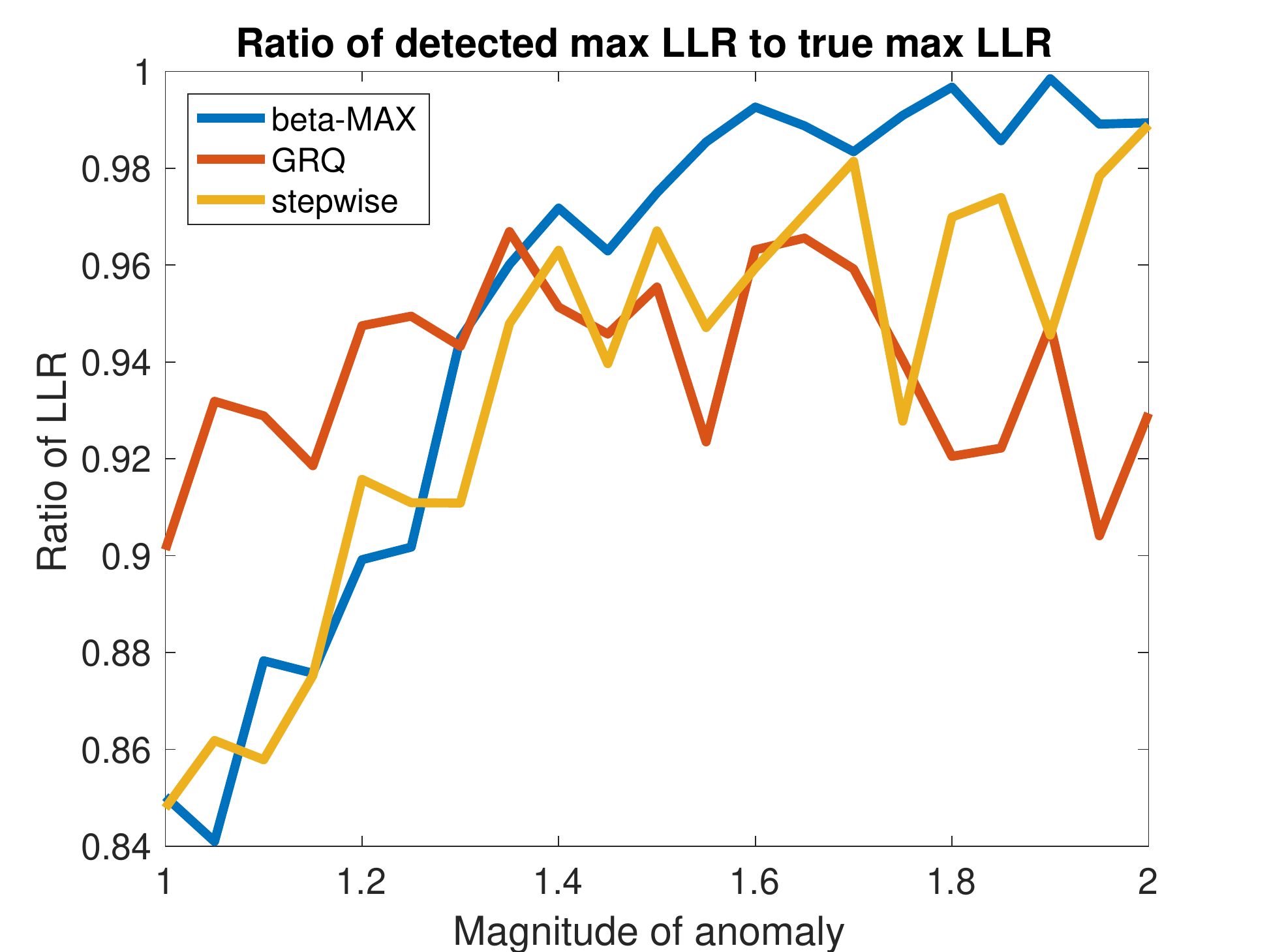}
  \end{minipage}%
  \begin{minipage}[b]{0.50\linewidth}
    \centering
    \includegraphics[width=1.1\linewidth]{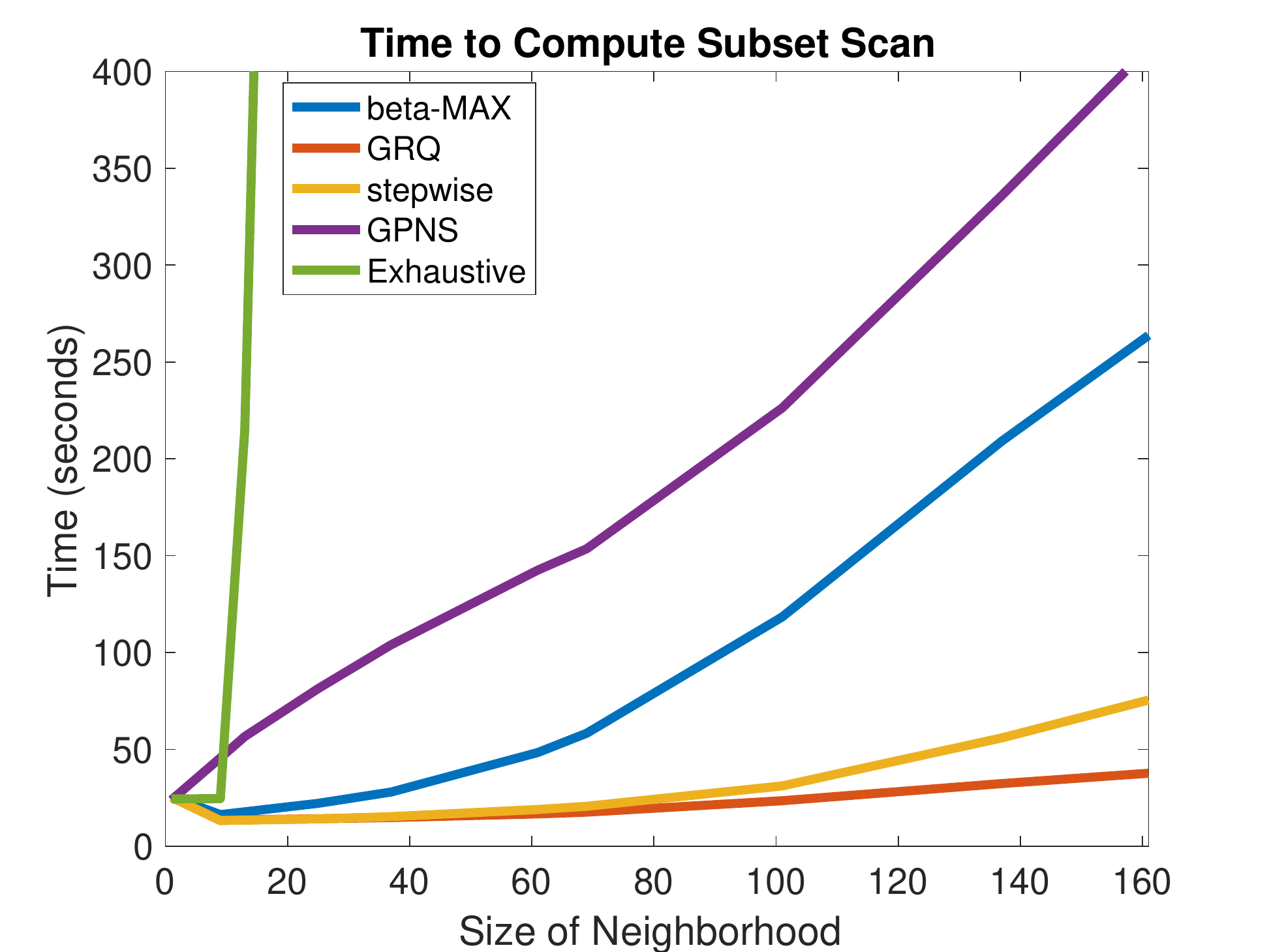}
\end{minipage}
\vspace*{-0.5cm} 
\caption{Numeric tests of GPNS and GPSS compared to exhaustive evaluation of $LLR(w^*)$. Left plot: ratio of maximum LLR identified by GPSS to true maximum LLR. Right plot: run time.}
\label{fig:LLR_and_time}
\end{figure}

Additionally, for each test we use an exhaustive search to find the subset with the highest LLR. The ratios of the LLR of approximate GPSS solutions to $LLR(w^*)$ are shown in Fig.~\ref{fig:LLR_and_time}. Note that all approximation methods are relatively close to the optimal value. While the $\beta_{MAX}$ approach dominates at large magnitudes, the GRQ dominates at small magnitudes and achieves a relatively stable ratio across all tests. 

To test the methods' scalability we vary the maximum neighborhood size and measure run time. In Fig.~\ref{fig:LLR_and_time} we compare GPSS, GPNS, and an exhaustive search for the optimal subset. The exhaustive search quickly becomes computationally intractable. Despite the added flexibility, GPSS is faster than GPNS because GP posterior inference is performed for fewer neighborhoods.

\begin{figure}[h]
\centering
\hspace*{-1.1cm} 
\includegraphics[width=1.25\linewidth]{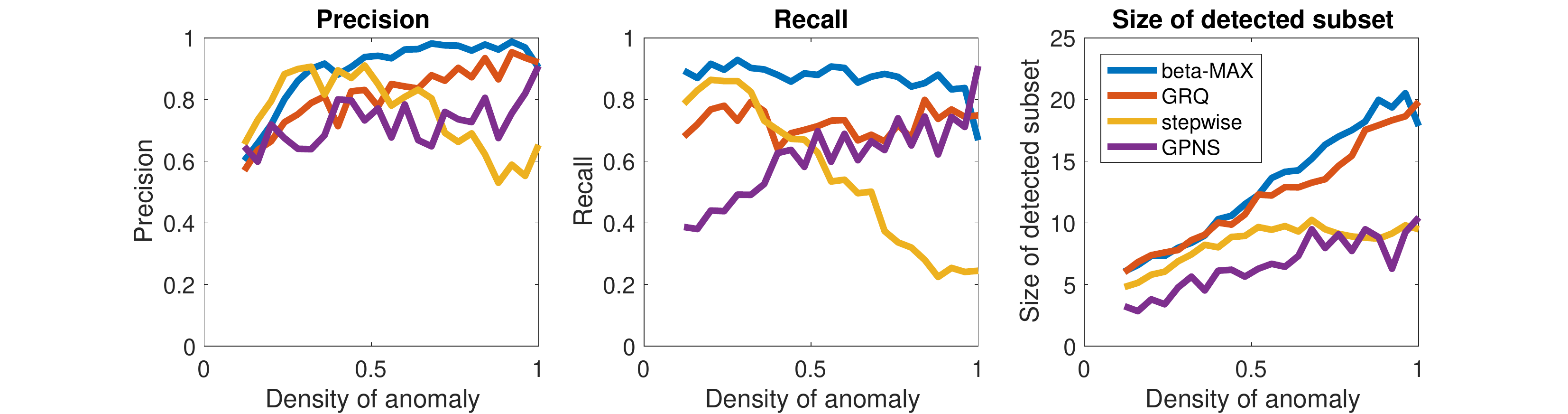}
\vspace{-0.4cm}
\caption{Precision, recall, and size of detected subset for GPSS and GPNS methods over subsets of varying density within a neighborhood.}
\label{fig:subsetDensity}
\end{figure}

We consider the effect of the density of anomalies on GPSS and GPNS where ``density'' is defined by the proportion of anomalous points in the true subset (Fig.~\ref{fig:subsetDensity}). 
While the stepwise method is competitive with the $\beta_{MAX}$ and GRQ approaches at low densities, its precision and recall drop off steeply at high densities. 
Additionally, in relatively low density anomalies, where the anomalous shapes may be highly irregular, GPNS has substantially reduced precision and recall.

\subsection{Urban opioid overdose deaths}
\label{sec:opioids}
\vspace{-1mm}
A recent United States opioid epidemic has garnered national attention~\citep{hhsopioid}. We study monthly opioid overdose deaths in New York from 1999-2015~\citep{CDCWONDER}. Data is provided at a county level for Manhattan, Brooklyn, Queens, the Bronx, Nassau County, and Suffolk County. Data is missing for some months in different counties. We apply GPSS and baseline approaches jointly to data across all time, latitude, and longitude, with randomization testing at $\alpha=0.05$.

All three GPSS approaches 
identify two statistically significant anomalous patterns. While precise points 

\begin{figure}[h!]
\centering
\subfloat{%
  \includegraphics[width=0.4\textwidth]{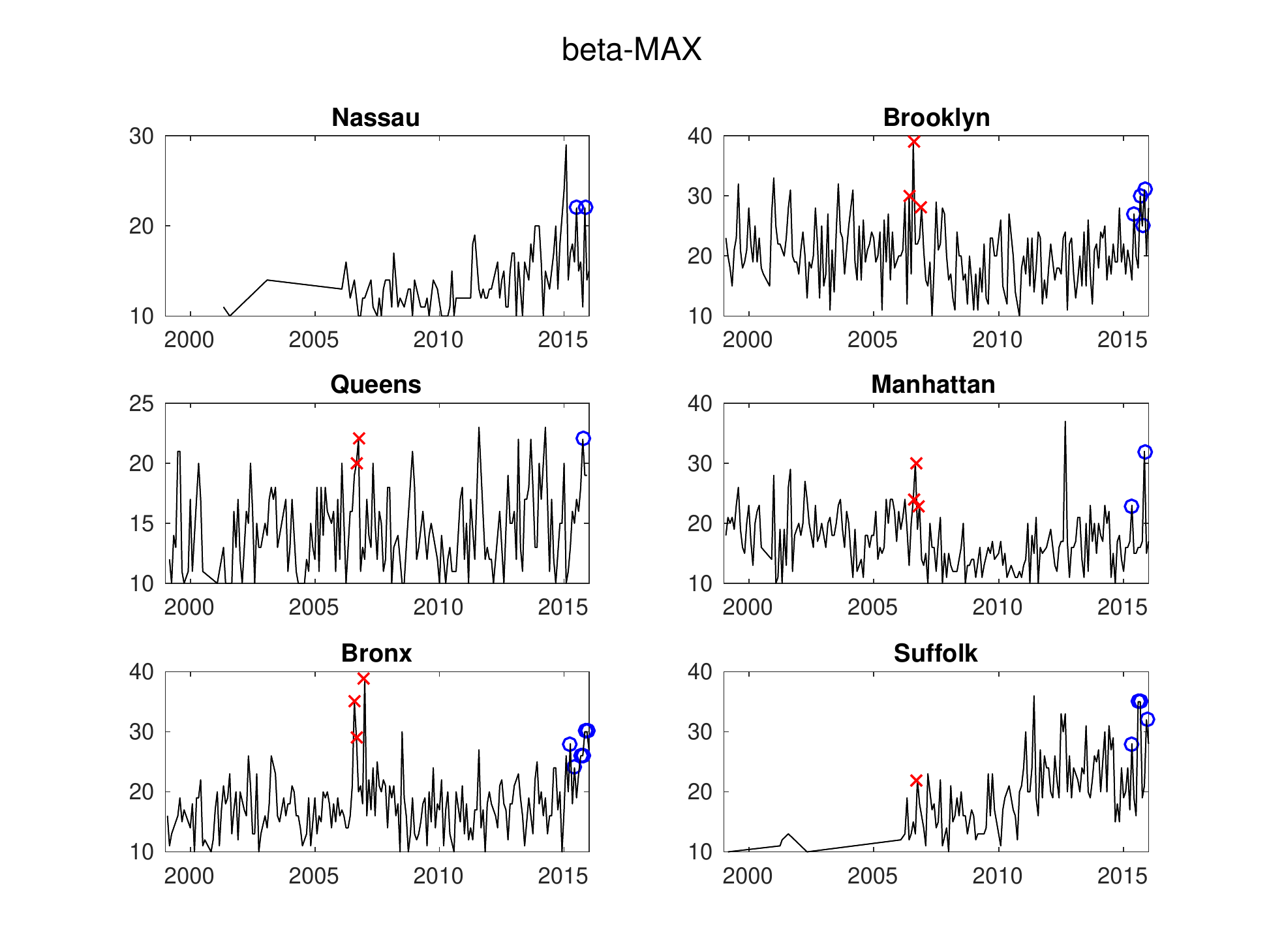}%
  }\par    
  \vspace*{-0.75cm}  
\subfloat{%
  \includegraphics[width=0.4\textwidth]{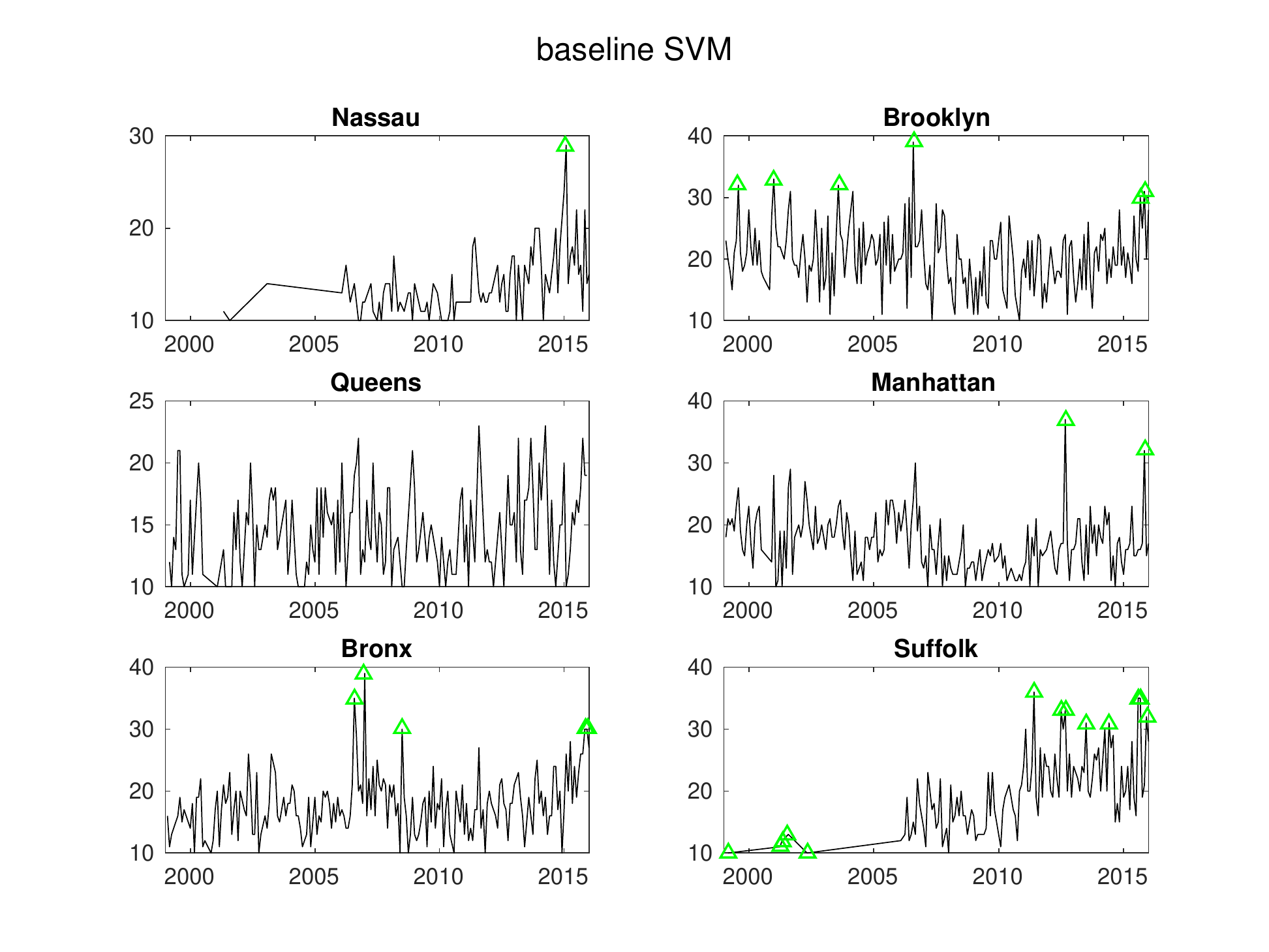}%
  }
  \vspace{-1mm}
\caption{Monthly opioid overdose deaths in New York from 1999-2015. Top plot depicts the two statistically significant anomalies detected by $\beta_{MAX}$. Bottom plot depicts points detected by the one-class SVM.}
\label{fig:opioid}
\end{figure}

selected by the methods differ slightly, Fig.~\ref{fig:opioid} depicts the two 
anomalous regions discovered by $\beta_{MAX}$ in blue circles and red crosses. With the exception of the independent subset scan, the baseline methods failed to discover a coherent anomalous pattern. Instead they selected individual points across space and time. For example, see results from the one-class SVM in Fig.~\ref{fig:opioid}. 


The anomalies detected by GPSS correspond to important public health events. The blue circles at the end of 2015 indicate a surge in opioid deaths corresponding to a well known plague of fentanyl-related deaths in NYC~\citep{HealingNYC}. The anomaly denoted by red crosses in 2006 is particularly interesting since it indicates a spike in opioid deaths immediately preceding the introduction of community training programs to administer a lifesaving naloxone drug. This may indicate a surge in fatalities that was cut short by making naloxone more widely available and educating communities in its use.

\vspace{-1mm}
\subsection{Manhattan 311 requests}
\label{sec:exp_311}
\vspace{-1mm}
New York City's 311 system enables residents to request government services. We consider a local public health event that occurred on 01/22/16 in upper Manhattan. On that day, local news reported that residents were concerned due to brown tap water~\citep{DNAwater,CBSwater}. Detecting the extent of the residents' concerns is important to help identify and mitigate public health risks.

\begin{figure}[h!]
\centering
\includegraphics[width=0.7\linewidth]{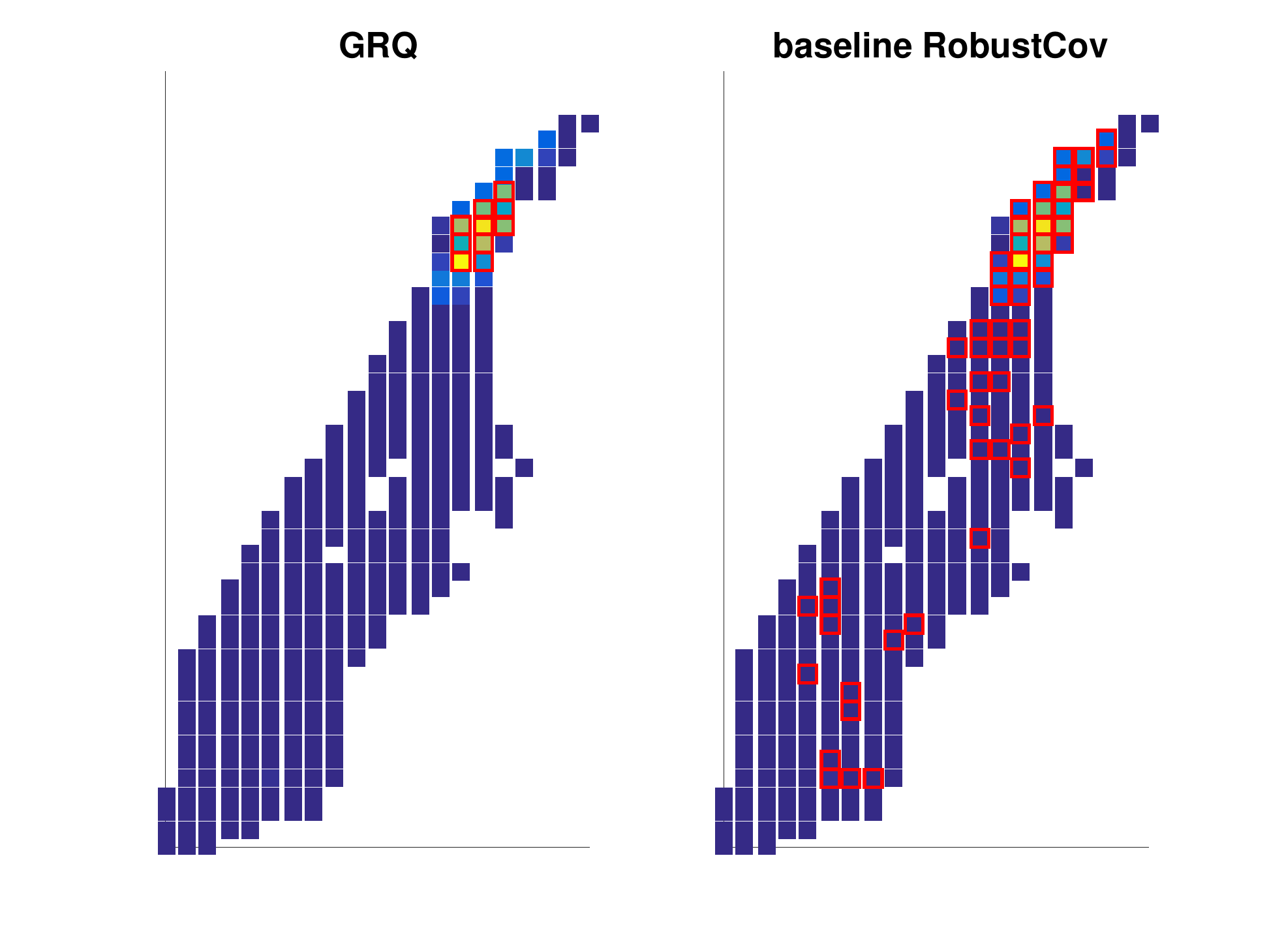}
\vspace{-4mm}
\caption{GPSS and robust covariance results for daily 311 requests in Manhattan on 01/22/16. Red squares indicate detected anomalies.}
\label{fig:311_water}
\end{figure}

We consider daily 311 requests in Manhattan for the month of January 2016, aggregated over a 0.08 mile$^2$ grid~\citep{NYCdata}. We apply GPSS methods and baseline approaches with neighborhoods of up to 15 points. All GPSS methods identified an anomalous pattern around the locations and time of the water discoloration event.
Baseline methods tended to substantially overestimate the anomaly's extent in both space and time. These results from January 22 are represented by the GRQ and the Robust baselines in Fig.~\ref{fig:311_water}. Blue and yellow squares indicate low and high volume of reports, respectively. Red squares indicate the top anomalous regions discovered by each method.

Ground truth does not exist for these hyper-local events so we cannot compute precision and recall. However, 311 requests have labeled types, although we used aggregated 311 calls as our data inputs. For each method we compute the ratio of water-related 311 calls to non-water-related calls in the detected anomalies. This ``water signal-to-noise'' ratio, listed in Table~\ref{tab:311_water}, indicates how precisely each method identified regions associated with many water-related requests. The entire dataset has a water signal-to-noise of 0.07.
\vspace{-2mm}
\begin{table}[h]
\centering
\caption{Signal-to-noise ratio of water-related 311 calls to non-water-related 311 calls for all methods.}
\label{tab:311_water}
\begin{tabular}{|l|l|}
\hline
\multicolumn{1}{|c|}{\textbf{Model}} & \multicolumn{1}{c|}{\textbf{Signal-to-Noise}} \\ \hline
GRQ                                  & 7.22                              \\ \hline
Stepwise                             & 7.22                              \\ \hline
$\beta_{MAX}$                        & 7.22                              \\ \hline
Independent SS                       & 7.06                              \\ \hline
Baseline GP                          & 0.44                              \\ \hline
One-class SVM                        & 0.23                              \\ \hline
RobustCov                            & 0.12                              \\ \hline
\end{tabular}
\end{table}
\vspace{-3mm}

\vspace{-1mm}
\subsection{Multi-stream: trees and sewers}
\label{sec:exp_tree_sewer}
\vspace{-1mm}

Using the multi-stream procedure from \S\ref{sec:mutli-stream}, we consider 311 reports of damaged trees and sewer issues. Both streams indicate weather-related issues: damaged trees indicate high winds while sewer calls indicate substantial precipitation. Together, these data identify areas with dangerous post-storm conditions. Each complaint type is fit with an independent GP and the entire data is scanned jointly for anomalies.

\begin{figure}[h]
\centering
\includegraphics[width=0.75\linewidth]{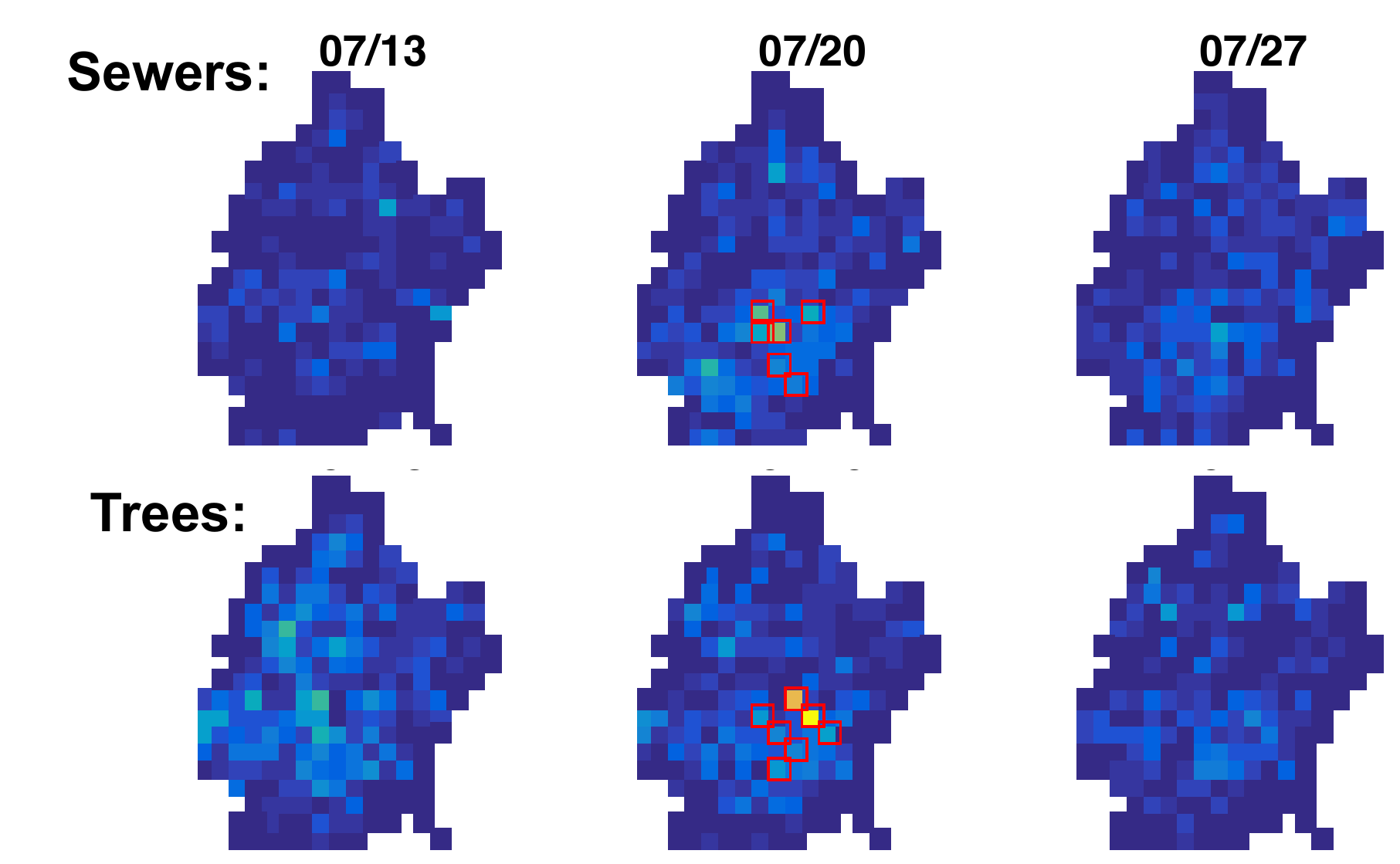}
\vspace{-0.2cm}
\caption{311 calls for damaged trees and sewer issues from 2016 in Brooklyn. Red squares indicate the top anomalies discovered by the $\beta_{max}$ approach.}
\label{fig:tree_sewer_2016}
\end{figure}
\vspace{-1mm}

\begin{figure}[h]
\centering
\includegraphics[width=0.75\linewidth]{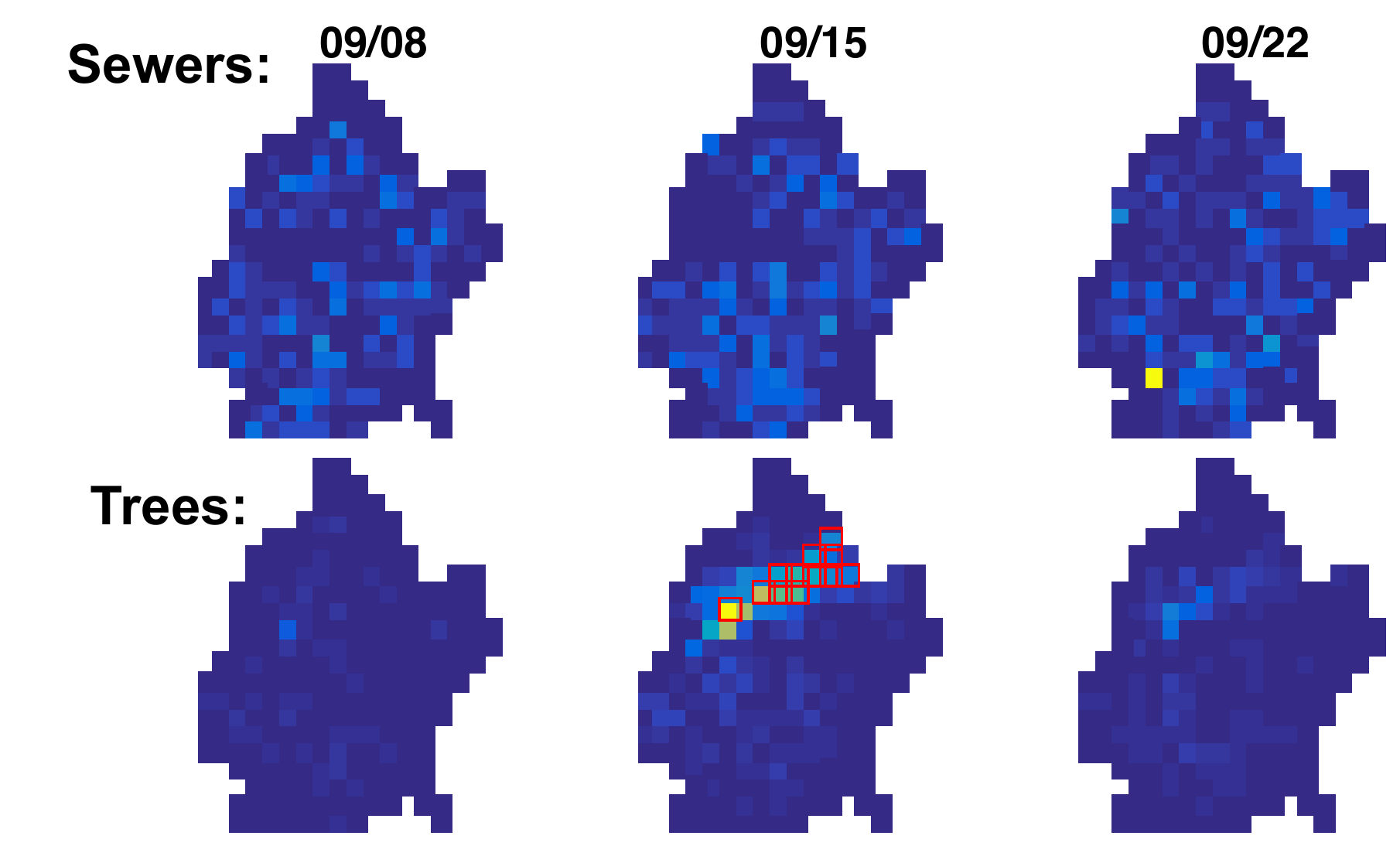}
\vspace{-0.2cm}
\caption{311 calls for damaged trees and sewer issues from 2010 in Brooklyn. Red squares indicate the top anomalies discovered by the $\beta_{max}$ approach.}
\label{fig:tree_sewer_2010}
\end{figure}

We analyze data in Brooklyn aggregated weekly over a 0.08 mile$^2$ grid~\citep{NYCdata}. We conduct analyses for 2016 and 2010 with results depicted in Figs.~\ref{fig:tree_sewer_2016} and ~\ref{fig:tree_sewer_2010}. The number of sewer reports (per week, per cell) are plotted on top, and damaged tree reports on bottom. 
Red squares indicate the top anomalous regions discovered using the $\beta_{max}$ approach. 

The most anomalous regions in 2016 were all concentrated during the week of July 20th when a significant summer storm felled trees and flooded sewers, thus jointly affecting both data streams~\citep{CBSstorm}.  Conversely, although the week of July 13th experienced elevated reports of felled trees no anomalous region is detected since there is no corresponding increase in sewer flooding. This demonstrates how multi-stream search may help to regulate GPSS.

\vspace{-1mm}

The most anomalous regions in 2010 were all concentrated during the week of September 15th when an urban tornado cut through Brooklyn~\citep{accuweatherTornado}. Unlike the 2016 results, these anomalies only occurred in reports of damaged trees. Also note the lone yellow square in the sewer data of September 22. Though the square indicates elevated number of calls, GPSS does not consider it anomalous since it does not represent a systematic shift in space and time.

\vspace{-1mm}
\section{Conclusions}
\label{sec:conclusions}
\vspace{-3mm}
We develop two GP-based subset scanning approaches to accurately and efficiently detect anomalous patterns in complex, highly correlated data. 
The results of GPNS and GPSS are coherent, powerful, and interpretable.  While the simpler GPNS method may be sufficient for circular clusters, GPSS provides additional flexibility to accurately identify irregular cluster shapes. Unlike individual anomaly detection methods, the spatial locality enforced by our methods ensures coherent explanation of detected anomalous patterns. As the 311 and opioid applications demonstrate, our approaches can be used for studying and informing policy decisions. Future work could  integrate categorical variables or extend to Student-t processes~\citep{shah2014student} for dealing with heavy-tailed noise.

\emph{This work was supported by NSF awards GRFP DGE-1252522 and  IIS-0953330, the RK Mellon Foundation, and NCSU Laboratory for Analytical Sciences.}

\appendix

\vspace{-1mm}
\section{Alternative model MLE}
\label{sec:appendix_MLE}
\vspace{-1mm}
Given data, $(x,y)$, we can determine the optimal mean shift, $\beta^*$ through maximum likelihood estimation as shown below. Let $\mu, \Sigma$ be the posterior mean and covariance of the null model in the domain of $x$, and denote $E = \Sigma^{-1}$ for brevity. 
\begin{equation}
\label{eq:MLE_GPSS}
\begin{aligned}
\beta^* = \max_\beta& \Big((2\pi)^{-\frac{k}{2}}|\Sigma|^{-\frac{1}{2}}\exp(-\frac{1}{2}(y-w\beta - \mu)^T \\
& E(y - w\beta - \mu))\Big) \\
= \max_\beta& -\frac{1}{2}(y-w\beta - \mu)^T E(y - w\beta - \mu) \\
= \max_\beta& (y - \mu)^T E w\beta  -  \frac{1}{2}(w\beta)^T E(w\beta)
\end{aligned}
\end{equation}
We take the derivative with respect to $\beta$ and set it to zero
\begin{equation}
\begin{aligned}
\frac{\delta LLR(w)}{\delta \beta} &= (y - \mu)^T E w  -  (w\beta^*)^T E(w) = 0 \\
&\Rightarrow  (w\beta^*)^T E(w) = (y - \mu)^T E w \\
&\Rightarrow  \beta^* = \dfrac{w^T E(y - \mu)}{w^T E w}
\end{aligned}
\end{equation}

\vspace{-1mm}
\section{Iterative $\beta_{MAX}$ algorithm to approximate optimal subset}
\label{sec:appendix_iterative}
\vspace{-1mm}
Since the derivation of $\beta_{MAX_i}$ is conditional on a subset $w$, we obtain the \textit{conditional} optimal subset. In order to approximate an optimal solution we use iteratively compute the conditional optimal subset beginning with a null subset, $w=\vec{0}$. This is an $O(\ell k \log(k))$ algorithm for some $\ell$ number of iterations, where $k$ is the size of the neighborhood. Pseudo-code is depicted in Alg.~\ref{alg:beta_max}.

\begin{algorithm}[h]
\SetAlgoLined
\KwResult{Highest scoring subset $w^*$ }
 Initialize $w=\vec{0}$\;
 \For{$l=1:\ell$}{
  Compute $\beta_{MAX_i}\: \forall i$ conditioned on the current value of $w$\;
  Find highest scoring subset, $w^{(l)}$, using a linear search over sorted $\beta_{MAX_i}$\;
  Compute $LLR(w^{(l)})$\;
  Set $w=w^{(l)}$\;
 }
 Choose $w^* = \argmax_{w^{(l)}} LLR(w^{(l)})$
 \caption{Iterative $\beta_{MAX_i}$ algorithm}
 \label{alg:beta_max}
\end{algorithm}

\vspace{-1mm}
\section{Constrained $\beta_{MAX}$ optimization over blocks} 
\vspace{-1mm}
Although we focus on unconstrained subsets searching within neighborhoods, real world applications sometimes require a more constrained optimization. For example, in spatiotemporal phenomena it is often useful to consider anomalous patterns that are nearby in space and contiguous over time. We can enforce such constraints by predefining mutually exclusive blocks of points, $(x^{(B)},y^{(B)}) \subseteq (x^{(n)},y^{(n)})$ where points in a block must all either be included in, or excluded from, a subset.

When considering blocks of points we can compute the total contribution from all points in the block, though we must also account for additional off-diagonal terms in $E$ due to the blocking of data points. Following the derivation in Section 4.2.1 of the main paper, we can derive the $\beta_{MAXb}$ for each block,
\begin{equation}
\begin{aligned}
\beta_{MAX_B} = \sum_{i\in B} \dfrac{ 2\big( E(y^{(n)}- \mu)\big)_i}{\big( \sum_{j\notin B} 2w_j E_{j,i} + E_{i,i} + \sum_{k\in B} E_{k,i} \big)}
\end{aligned}
\end{equation}
This can be used in a lightly modified version of Algorithm~\ref{alg:beta_max} where the $\beta_{MAXb}$ of blocks, not individual points, is iteratively computed.

\vspace{-1mm}
\section{School Absenteeism}
\label{sec:exp_school}
\vspace{-1mm}
Public schools in New York City record and publish daily student attendance~\citep{NYCdoe}. Given the importance of education on future outcomes there is tremendous interest in understanding patterns of school absenteeism. We consider public school attendance data in Manhattan for the 2015-2016 school year. The data is messy, with missing entries and non-uniform placement of school locations. We aggregate data at weekly level and remove the last four weeks of the school year since they contain known high absenteeism rates that are not of interest to Department of Education officials. 

\begin{figure}[h]
\centering
\includegraphics[width=0.8\linewidth]{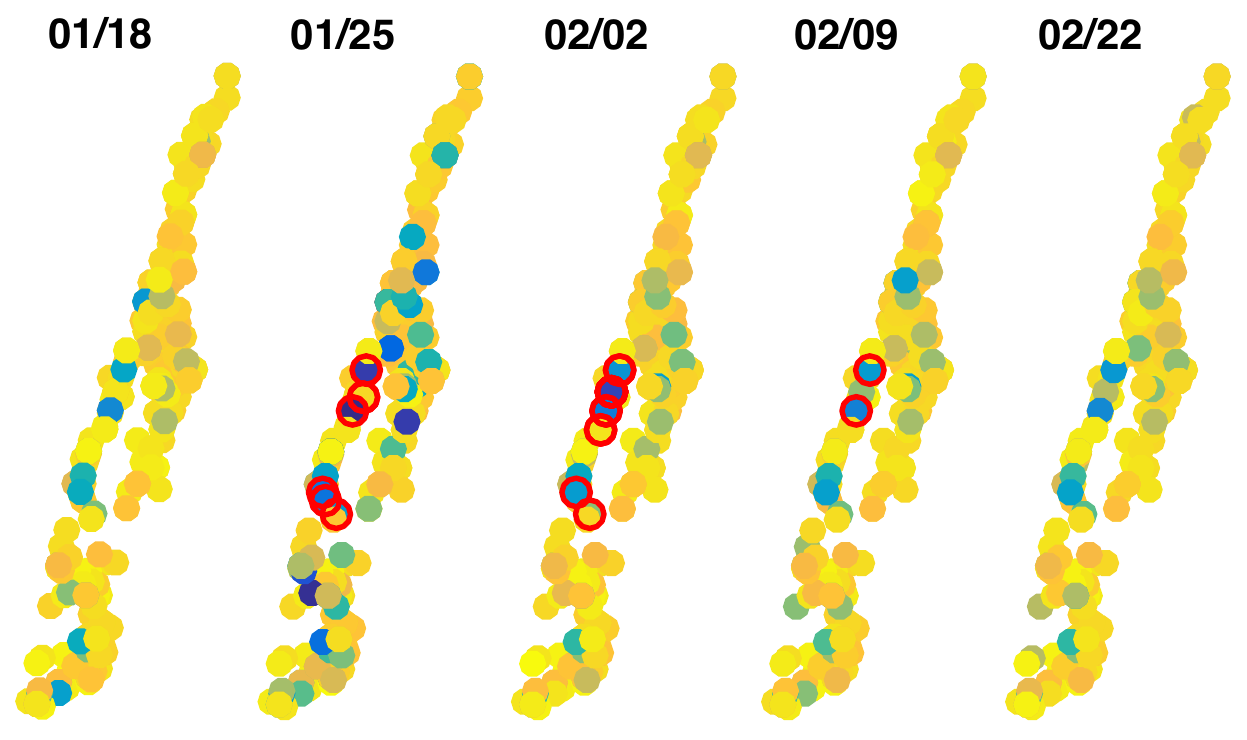}
\vspace{-0.4cm}
\caption{School absenteeism results from Manhattan using GRQ. Each dot represents a school location, with yellow dots indicating high attendance and blue dots indicating low attendance. The space-time locations of schools in the top ten anomalous subsets are bordered in red.}
\label{fig:school_man_red}
\end{figure}

We apply GPSS methods and baseline approaches with neighborhoods of up to ten local schools. All GPSS methods identified an anomaly around January to February 2016 concentrated on West Side of Manhattan.  The results from GRQ around the time of the detected anomaly are presented in Fig.~\ref{fig:school_man_red}. Each dot represents a school location, with yellow dots indicating high attendance and blue dots indicating low attendance. The space-time locations of schools in the top ten anomalous subsets are bordered in red.

The detected anomalies correspond to a category five blizzard which may have disrupted teachers and students from attending school even though no snow day closings were reported at the time. Further research is required to understand why the West Side of Manhattan differed systematically from the rest of the borough. Baseline anomaly detection methods did not identify a coherent anomaly and instead detected anomalies throughout the year.

\small
\bibliographystyle{plainnat}
\bibliography{herlands_aistats}

\end{document}